\documentclass{article}
\usepackage{graphicx} 
\graphicspath{ {./images/} }
\usepackage{float}
\restylefloat{table}
\usepackage{arxiv}
\usepackage{hyperref}
\hypersetup{
    colorlinks=true,
    linkcolor=blue,
    filecolor=blue,      
    urlcolor=blue,
    citecolor = blue,
}
\usepackage[justification=centering]{caption}
\usepackage[utf8]{inputenc} 
\usepackage[T1]{fontenc}    
\usepackage{hyperref}       
\usepackage{url}            
\usepackage{booktabs}       
\usepackage{amsfonts}       
\usepackage{nicefrac}       
\usepackage{microtype}      
\usepackage{lipsum}
\usepackage{comment}
\usepackage{caption,setspace}
\usepackage{natbib}
\usepackage{amssymb} 
\usepackage{amsmath}
\title{Feature Selection and Extraction for Graph Neural Networks}
\setcitestyle{round,semicolon,open={(},close={)}}

\author{
 Deepak Bhaskar Acharya\\
  Computer Science Department\\
  University of Alabama in Huntsville\\
  \texttt{deepak.acharya@uah.edu} \\
  \texttt{deepakacharyab@gmail.com} \\
   \And
 Dr. Huaming Zhang \\
   Computer Science Department\\
  University of Alabama in Huntsville\\
  \texttt{zhangh2@uah.edu} \\
  \\
}

\begin{document}
\maketitle

\begin{abstract}


Graph Neural Networks (GNNs) have been a latest hot research topic in data science, due to the fact that they use the ubiquitous data structure graphs as the underlying elements for constructing and training neural networks. In a GNN, each node has numerous features associated with it. The entire task (for example, classification, or clustering) utilizes the features of the nodes to make decisions, at node level or graph level. In this paper, (1) we extend the feature selection algorithm presented in  via Gumbel Softmax to GNNs. We conduct a series of experiments on our feature selection algorithms, using various benchmark datasets: Cora, Citeseer and Pubmed. (2) We implement a mechanism to rank the extracted features. We demonstrate the effectiveness of our algorithms, for both feature selection and ranking. For the Cora dataset, (1) we use the algorithm to select 225 features out of 1433 features. Our experimental results demonstrate their effectiveness for the same classification problem. (2) We extract features such that they are linear combinations of the original features, where the coefficients for each extracted features are non-negative and sum up to one. We propose an algorithm to rank the extracted features in the sense that when using them for the same classification problem, the accuracy goes down gradually for the extracted features within the rank 1 - 50, 51 - 100, 100 - 150, and 151 - 200. 

\end{abstract}

\keywords{Graph Neural Networks \and Gumbel-Softmax \and Feature Selection \and Feature Extraction}

\section{Introduction}
One of the common problems in machine learning is the curse of high dimensionality. The amount of training data required could grow exponentially along with the dimension of the data \citep{Bauer}. Thus, feature selection, which aims to reduce the high dimensionality by identifying and selecting the subset of most relevant features in a dataset has been one of the main research topics for data science. There are many downstream benefits for feature selection: (1) the data mining algorithms using the selected features can be faster; (2) the accuracy of the trained model can be improved; (3) the model can be more interpretable; and (4) the overfitting problem can be alleviated by removing irrelevant or redundant features.

 Feature selection is different from the more general problem of dimensionality reduction. Standard techniques for the dimensionality reduction such as Principle Component Analysis (PCA)  \citep{Kambhatla} and AutoEncoder \citep{baldi2012autoencoders} reduces the dimensionality by constructing a new set of attributes from the existing set of features and extracting from those newly constructed attributes. The goal is to preserve the maximal variance (for PCA) or to minimize the reconstruction loss (for AutoEncoder). None of them selects existing features from the given dataset. Thus, none of them can be used to eliminate redundant or irrelevant features for a specific downstream supervised task such as classification. In addition, the newly constructed attributes are difficult to be interpreted, which makes them unfeasible for designing interpretable machine learning algorithms.

Very recently, \citeauthor{abid2019concrete} (\citeyear{abid2019concrete}) proposed concrete AutoEncoder for feature subset selection. Concrete AutoEncoder uses a differentiable relaxation of the concrete distribution \citep{maddison2016concrete}, and the reparameterization trick \citep{kingma2013auto} to differentiate through a loss function (for example, the reconstruction loss) and to select input features to minimize the loss. Applying their method, they selected 20 features out of 784 features for the MNIST dataset. They were able to reconstruct the images with high accuracy using selected 20 features out of 784 features by using a deep AutoEncoder.



GNN is an extension of existing neural network for processing the data in the graph domains \citep{zhou2018graph,wu2019comprehensive,Scarselli}. Graphs are one of the data structures consisting of nodes and edges in which some pairs of the nodes are in some sense "related" through edges. Graph analysis with the machine learning technique have been recognized as the power of graphs is immense \citep{Keyulu} i.e., graphs can be used as denotation of a large number of systems across various areas including social networks \citep{Hamilton,Welling}, knowledge graphs \citep{Hamaguchi}, natural science (physical systems \citep{SanchezGonzalez, BattagliaPascanu} and protein-protein interaction networks \citep{Shariat}) and many other research areas \citep{Khalil}. 

In this paper, (1) we extend the feature selection algorithm presented by \cite{abid2019concrete} via Gumbel Softmax to GNNs. We conduct a series experiments, using a variety of benchmark datasets: Cora, Citeseer and Pubmed. (2) We design a mechanism to rank the selected features. We demonstrate the effectiveness of our algorithms, for both feature selection and ranking. As an illustrating example for the Cora dataset, we select 225 features out of 1433 features and we rank them according to our mechanism. The experiments results show that, the accuracy goes down gradually when using the selected features falling in the range of rank 1 - 50, 51 - 100, 100 - 150, and 151 - 200 for the same classification problem\footnote{Code available at: \url{https://github.com/deepakacharyab/gnn_feature_selection_extraction.git}.}.  


The remaining of the paper is organized as follows. In Section \ref{Related work}, we introduce background and related works in more details. In Section \ref{Proposed Method}, we introduce our main method. In Section \ref{Experiment Results}, we present the experimental results.

\section{Background and Related work}\label{Related work}

There are three common approaches for feature selection: filter, embedded and wrapper \citep{shardlow2016analysis}. In filter approach, features are pre-determined by using some algorithms and then the selected ones are used in machine learning algorithm. For example, one can use Pearson’s correlation coefficient to find features which are more relevant to the class label. The higher the value of Pearson's correlation coefficient, the more relevant the feature is to the class label. Hence, they are more useful for any downstream machine learning tasks. In embedded approach, the machine learning algorithm decides the relevant features to be used or ignored, such as decision trees. In wrapper approach, feature subset selection is done as a black box where no knowledge on how the algorithm searches and finds a good subset the features. \citep{kohavi1997wrappers}.

Recently, Gumbel Softmax was successfully used for feature selection.
Gumbel-Softmax distribution is "a continuous distribution over the simplex that can approximate samples from a categorical distribution" \citep{Shixiang}. A categorical distribution, can be viewed as a one-hot vector by identifying the maximum probability to one, and all the other probability to zero. The one-hot vector can be interpreted as selecting one feature from all the features (its meaning will be clearer later), where one in the vector indicates the corresponding feature is selected while zero in the vector indicates the corresponding feature is not selected. 

Let z be a categorical random variable with class probabilities $\pi_1, \pi_2, \cdots ,\pi_k$. The Gumbel-Max trick \citep{Julius,maddison2014sampling} provides a simple and effective way to draw samples z from a categorical distribution with the designated class probabilities $\pi$:
\begin{equation} \label{argmax}
z = one\_hot (arg max_i[g_i + log \pi_i])
\end{equation}

where $g_1,g_2,\cdots,g_k$ are i.i.d samples drawn from the Gumbel(0,1), which can be sampled via inverse transform sampling as follows: $g_i = -log(-log(u))$, $u \sim$ Uniform$(0,1)$. 

\begin{figure}[!htbp]
    \centering
    \begin{minipage}{0.3\textwidth}
      \includegraphics[scale=0.5]{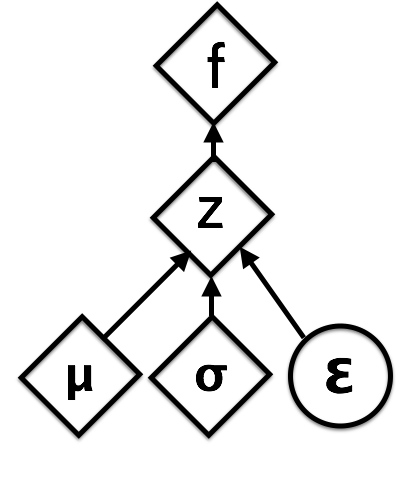}
    \end{minipage}\hfill
    
    \caption{Re-parameterization trick} 
    \label{repara_img}
\end{figure}

The training of a neural network via gradient descent requires every operation in the network being differentiable. Note that, in Equation \ref{argmax}, the argmax function and the stochastic sampling operations $g_i$, where $i = 1,2,\cdots,k$ are not differentiable. Firstly, a typical solution for making argmax differentiable is to approximate it by a softmax function. Furthermore, one can use a temperature $\tau$ to control the level of approximation for argmax as follows:     

\begin{equation} \label{actual_gumbel}
y_i=\frac{exp((log(\pi_i)\ +\ g_i)/\tau)}{\sum_{j=0}^{k}{exp((log(\pi_j)\ +\ g_j)/\tau)}} \ for \ i = 1,2,3, ...... ,k.
\end{equation}

In Equation \ref{actual_gumbel}, one can easily see that as the temperature $\tau$ $\rightarrow 0$,  samples from the above Gumbel Softmax distribution (as introduced in \citep{Shixiang,maddison2016concrete}) become one-hot and the Gumbel-Softmax distribution becomes identical to the
categorical distribution $p(z)$. Secondly, the indifferentiability of the stochastic operations $g_i$, where $i = 1,2,\cdots,k$ can be solved by using the reparametrization trick (\cite{Shixiang}). Essentially, the reparametrization trick rewrites each stochastic operation into a differentiable operation of its non-stochastic parameters and a random variable with a fixed distribution, where a random sampling output from the fixed distribution is treated as an additional input to the operation. By doing so, the stochasticity of the stochastic operation has no prohibiting effect on backpropagating the gradients. Thus, gradient descent algorithm is still applicable and the entire neural network can still be trained. For example, consider the stochastic operation $z \sim \mathcal{N} (\mu,\sigma)$. It can be re-written as z = $\mu + \sigma \cdot \epsilon$, where $\epsilon \sim \mathcal{N} (0,1)$. Once treating $\epsilon$ as an input, the gradient with respect to $z$ can be backpropagated to $\mu$ and $\sigma$, which can further be backpropagated.
 The operation node $z$ is stochastic, where backpropagation can not go through. In Figure \ref{repara_img}\footnote{Figure taken from: \url{https://towardsdatascience.com/generating-images-with-autoencoders-77fd3a8dd368}.}, the stochastic operation node becomes deterministic, since the sampling is done with the $\epsilon$ node, whose sampling output is treated as an ordinary input to $z$. Hence, backpropagation in the operation node $z$ still goes through. Applying the reparametrization trick, each stochastic operation $g_i$, where $i = 1,2,\cdots,k$ in Equation \ref{actual_gumbel} becomes a deterministic function of an input (which is sampled from the fixed Gumbel(0,1) distribution). Thus, the gradient with respect to $\pi_i$, where $i= 1,2,\cdots,k$ can be estimated with low variance through backpropagation.

Gumbel-softmax was used in the concrete AutoEncoder for feature selection \citep{abid2019concrete}. In contrast to AutoEncoder, which uses fully-connected layers for the encoder, concrete Autoencoder uses a single feature selection layer to select the features. It uses discrete categorical distribution to approximate one-hot vectors for feature selection, where the discrete categorical distribution is approxiamted by the Gumbel Softmax distribution, when the $\tau$ $\rightarrow 0$, the Gumbel Softmax is identical to the catogerical distribution. The exprimental results on MNIST dataset and the other data sets demonstrated that they are quite accurate. 




\section{Proposed Method}\label{Proposed Method}

Inspired by the success of concrete AutoEncoder for feature selection \citep{abid2019concrete} and the concept of Gumbel-Softmax we tried to select features of a GNN with the citation network datasets such as Cora, Citeseer and Pubmed. 

Consider the graph with 'n' nodes and 'f' features then by applying the concept of feature selection we bring down the number of features from $f$ to $k$ where $k$ indicates the number of features selected (where $k < f$). Then, we use the gumbel feature selection matrix (i.e., the matrix which has the features selected when Gumbel-Softmax applied) to train the dataset and test how the accuracy of the selected features work. Our proposed deep learning model performs good with reduced features which is around 80-85\% reduction in the number of features initially the dataset had. 

In general, let $X_{n \times f}$ be the input feature matrix where, 'n' represents the total number of nodes and 'f' represents the total number of features for each node in the graph dataset. Consider $M_{f \times k}$ where, 'f' represents the total number of features in the graph dataset and 'k' represents the features we are selecting out of 'f' features. This method can be used for either feature selection or feature extraction.  

{\bf Feature Selection:}
From the Equation \ref{actual_gumbel}, as the temperature $\tau \rightarrow 0$, every column vector sampled from the Gumbel-Softmax distribution obtained from its corresponding column vector in the matrix ${\cal W}$ is a categorical distribution. After setting the largest probability value in the column to $1$ and remaining values to $0$, then the column vector becomes a one-hot vector. We denote the resulting matrix still by ${\cal W}$. Consider the feature matrix $X_{n \times f}$. The product matrix $X_{n \times f} \times  {\cal W}_{f \times k}$ can be viewed as selecting $k$ features from the original feature matrix ${X}$. Totally, we selected 'k' features out of 'f' features, since ${\cal W}_{f \times k}$ has $k$ columns.

{\bf Feature Extraction:}
From the Equation \ref{actual_gumbel}, for any temperature $\tau > 0$, every column vector sampled from the Gumbel-Softmax distribution obtained from its corresponding column vector in 
the matrix $\mathbb{W}$ is a categorical distribution. Then, the product matrix $X_{n \times f} \times \mathbb{W}_{f \times k}$ can easily be interpreted as extracting features from the original feature matrix ${X}$, where each extracted feature is a linear combination of the original features and the coefficients for it are non-negative numbers and sum up to one. 

We used negative log likelihood loss (NLLLoss) as a loss function to calculate the loss and Adam optimizer as a metric for optimization.

The two layer Graph Convolution Network (GCN) used in our experiment is defined as 
 
\begin{equation} \label{gcn_1}
GCN(X,A) = Softmax(A (ReLu(AXW_GW_1)) W_2) 
\end{equation}

To verify the selected features and calculate the accuracy for classification we use the following two layer Graph Convolution Network as defined below

\begin{equation} \label{gcn_2}
GCN(X,A) = Softmax(A (ReLu(AXW_G')) W_2) 
\end{equation}

$A$: Adjacency matrix of the undirected graph G. 

$X$: Input feature matrix. 

$W_G$: Gumbel-Softmax feature selection / feature extraction matrix. 

$W_G'$: feature selection / feature extraction matrix obtained from the result of Equation \ref{gcn_1}.

$W1,W2$: Layer-specific trainable weight matrix.

$ReLu$ : Activation function ReLu(.) = max(0,.).

\section{Experiment Results}\label{Experiment Results}
    \begin{table}
    \label{tab:table}
    \end{table}
    \subsection{Datasets}
    \begin{flushleft}
    
    To test and verify the effectiveness of our experiment on the feature selection we use semi-supervised classification benchmarking datasets such as Cora, Citeseer and Pubmed which are standard citation network. In all of these datasets, nodes correspond to documents and edges to (undirected) citations. Node features correspond to elements of a bag-of-words representation of a document. 
    
    The Cora dataset contains 2708 papers from 7 classes such as Reinforcement Learning, Neural Networks, Case Based, Genetic Algorithms, Probabilistic Methods, Rule Learning, and Theory. The Citeseer dataset contains 3327 papers from 6 classes such as HCI, DB, Agents, AI, ML, and IR.
    
    Table \ref{tab:information about the datasets used in our experiment}. shows the list of datasets used in the experiments. Each dataset has the nodes, edges, features and the number of classes associated with them. The predictive power of the trained models is evaluated on 1000 test nodes, and we use 300 nodes for validation purposes.
    
    \end{flushleft}
    \begin{table}[ht]
    \centering
    \begin{tabular}{|l|c|c|c|c|c|}
        \midrule
        \cmidrule(r){1-6}
        Dataset     & Nodes & Edges & Features &Classes  &Accuracy (In \%)\\
        \midrule
         Cora     &2,708 &5,429  &1,433 &7 &83.50\\
         \midrule
         Citeseer &3,327 &4,372  &3,703 &6  &73.00    \\
         \midrule
         Pubmed   &19,717 &44,338 &500 &3 &80.00\\
        \midrule
    \end{tabular}
    \caption{Information about the datasets used in our experiment with current benchmarking accuracy, trained on all features.}
    \label{tab:information about the datasets used in our experiment}
    \end{table}%
    \subsection{Results}
    We conduct our experiments in two steps. In the first step we train gumbel feature selection matrix for fixed number of epochs which gives us the features selection matrix and in the second step we use trained gumbel feature selected matrix which has the prominent and dominating  features to classify the nodes and verify the accuracy of the model. 
    
    Table \ref{tab:Benchmark performance for node classification} shows the benchmark performance for node classification on three frequently used data sets.  Cora, Citeseer, and Pubmed are evaluated by classification accuracy. The listed methods use the same train/valid/test data split. Table \ref{tab:feature selection accuracy} gives information of the accuracy for the feature selection.  
    In Table \ref{tab:Test Results for Cora dataset based upon the ranking}, 
    Table \ref{tab:Test Results for Citeseer dataset based upon the ranking}, and
    Table \ref{tab:Test Results for Pubmed dataset based upon the ranking}, 
    the first column indicates the extracted features ranked according to their prominence in the dataset and the second column indicates the accuracy of the model. For example, consider Table \ref{tab:Test Results for Cora dataset based upon the ranking}, 
    the first row indicates the rank 1 to rank 225 prominent features out of 1,433 features of Cora dataset which gives the accuracy of 73.80\% over the highest accuracy of 83.50\% (from Table \ref{tab:Benchmark performance for node classification}) when all 1,433 features were present.

    \begin{table}[!htbp]
    \centering
    \begin{tabular}{|l|c|c|c|}
        \midrule
        \cmidrule(r){1-4}
        Method     & Cora & Citeseer & Pubmed \\
        \midrule
         GCN (2016) \citep{Welling}    &81.50 &70.30  &79.00 \\
         \midrule
         DualGCN (2018) \citep{Qiang} &83.50 &72.60  &80.00       \\
         \midrule
         GAT (2017) \citep{Petar}   &83.00 &72.50 &79.00  \\
         \midrule
         LGCN (2018) \citep{Hongyang} &83.30 &73.00 &79.50 \\
         \midrule
         StoGCN (2017) \citep{Stochastic} &82.00 &70.90 &78.70\\
         \midrule
         DGI (2018) \citep{Fedus} &82.30 &71.80 &76.80 \\
         \midrule
    \end{tabular}
    \caption{Benchmark performance for node classification}
    \label{tab:Benchmark performance for node classification}
    \end{table}%

    \begin{table}[!htbp]
    \centering
    \begin{tabular}{|l|c|c|c|}
        \midrule
        \cmidrule(r){1-2}
        Dataset &Selected Features     & Accuracy ( In \% )\\
        \midrule
        Cora &225  &68.20\\
        \midrule
        Citeseer &450    &57.70\\
         \midrule
        Pubmed &105  &66.80 \\
        
         \midrule
    \end{tabular}
    \caption{Classification accuracy results, trained on our selected features.}
    \label{tab:feature selection accuracy}
    \end{table}%

   \begin{table}[!htbp]
    \centering
    \begin{tabular}{|c|c|}
        \midrule
        \cmidrule(r){1-2}
        Extracted Features (Ranking)     & Accuracy (50 epochs) \\
        \midrule
        1 - 225  &73.80\\
        \midrule
        \midrule
         1 - 50    &61.30 \\
         \midrule
         50 - 100  &54.80       \\
         \midrule
         100 - 150   &48.80  \\
         \midrule
         \midrule
         1 - 75  &66.10  \\
         \midrule
         75 - 150   &58.40 \\
         \midrule
         150 - 225  &54.40  \\
         \midrule
    \end{tabular}
    \caption{Classification accuracy results, feature extraction for Cora dataset based upon the ranking}
    \label{tab:Test Results for Cora dataset based upon the ranking}
    \end{table}%
    
    \begin{table}[!htbp]
    \centering
    \begin{tabular}{|c|c|}
        \midrule
        \cmidrule(r){1-2}
        Extracted Features (Ranking)     & Accuracy (50 epochs) \\
        \midrule
        1 - 450   &61.80\\
        \midrule
        \midrule
         1 - 75    &52.00 \\
         \midrule
         75 - 150  &50.30       \\
         \midrule
         150 - 225   &48.90  \\
         \midrule
          225 - 300   &50.50  \\
         \midrule
          300 - 375   &44.90  \\
         \midrule
         \midrule
         1 - 150  &58.00  \\
         \midrule
         150 - 300   &58.00 \\
         \midrule
         300 - 450  &51.9  \\
         \midrule
    \end{tabular}
    \caption{Classification accuracy results, feature extraction for Citeseer dataset based upon the ranking}
    \label{tab:Test Results for Citeseer dataset based upon the ranking}
    \end{table}%
    
    \begin{table}[!htbp]
    \centering
    \begin{tabular}{|c|c|}
        \midrule
        \cmidrule(r){1-2}
        Extracted Features (Ranking)     & Accuracy (100 epochs) \\
         \midrule
         1 - 105   &71.10  \\
        \midrule
        \midrule
         1 - 35    &67.10 \\
         \midrule
         35 - 70  &60.00       \\
         \midrule
         70 - 105   &54.60  \\
         \midrule
        
    \end{tabular}
    \caption{Classification accuracy results, feature extraction for Pubmed dataset based upon the ranking}
    \label{tab:Test Results for Pubmed dataset based upon the ranking}
    \end{table}%

\section{Conclusion}
We introduced the feature selection and feature extraction method for the GNNs. The experimental results demonstrates the effectiveness of both the methods. In particular, for feature selection we use 15-20\% of the total features and still get the accuracy of 80-85\% of the benchmarking results (where all the features are used). Based on the experimental results, the ranking algorithm for the features extracted performs well on classification problem. We are still working on the ranking of feature selection method.


\bibliographystyle{apalike}
\bibliography{references}
\clearpage
\end{document}